# Transformed Subspace Clustering

**Jyoti Maggu**[1], **Angshul Majumdar**[1] and **Emilie Chouzenoux**[2]
[1]Indraprastha Institute of Information Technology
email: {jyotim, angshul}@iiitd.ac.in
[2]CentraleSupelec
email: emilie.chouzenoux@centralesupelec.fr

**Abstract**

Subspace clustering assumes that the data is separable into separate subspaces. Such a simple assumption, does not always hold. We assume that, even if the raw data is not separable into subspaces, one can learn a representation (transform coefficients) such that the learnt representation is separable into subspaces. To achieve the intended goal, we embed subspace clustering techniques (locally linear manifold clustering, sparse subspace clustering and low rank representation) into transform learning. The entire formulation is jointly learnt; giving rise to a new class of methods called transformed subspace clustering (TSC). In order to account for non-linearity, kernelized extensions of TSC are also proposed. To test the performance of the proposed techniques, benchmarking is performed on image clustering and document clustering datasets. Comparison with state-of-the-art clustering techniques shows that our formulation improves upon them.

## 1 Introduction

The problem of clustering is well known. It studies how signals are naturally grouped together. Perhaps the simplest and most widely used clustering technique is the K-means [1]. It groups the samples such that the total distance of the data points within the cluster are minimized. The problem is NP hard, and hence is usually solved greedily.

One of the limitations of K-means is that it operates on the raw data and hence fails to capture non-linear relationships. The simple fix to that is the kernel K-means [2]. The concept remains the same as in any kernel trick; operationally instead of Euclidean distances between the samples, its kernelized version is used for K-means.

Related to the kernel K-means is spectral clustering [2, 3]. The kernelized data matrix is the same as the affinity matrix for spectral clustering. It just generalizes the kernels to be any similarity measure and not necessarily based on Mercer kernels.

Subspace clustering techniques [4] assume that the samples from the same cluster will lie in the same subspace. Operationally, it involves expressing each data point as a linear combination of other data points. These linear weights serve as inputs for creating the affinity matrix. The factorization based techniques prevalent in clustering (such as [5-7]) technically belong to the subspace clustering paradigm.

In the past, it has been found that instead of applying subspace clustering on the raw data, a projection can be learnt such that the clustering is carried out in the projected domain. In [8, 9] a tight-frame was learnt from the data along with the subspace clustering formulation. The fundamental assumption behind such formulations is that even if the original data do not fall on separate sub-spaces, their projected versions will.

Our work is based on similar assumptions. Instead of applying subspace clustering on the original space, we will learn the subspace clustering in the transformed space. Therefore, even if the data cannot be segmented / clustered in the original domain, its transformed representation can be clustered into separate subspaces. The work proposes to incorporate three variants of subspace clustering – i) Locally linear manifold clustering (LLMC), ii) sparse subspace clustering (SSC), and iii) low rank representation (LRR).

We compared our proposed formulations with state-of-the-art representation learning and clustering techniques. We show that our method improves over the rest by a considerable margin in terms the clustering metrics used here.

## 2 Literature Review

### 2.1 Subspace Clustering

Subspace clustering techniques like locally linear manifold clustering (LLMC) [11, 12], sparse subspace clustering (SSC) [13-15] and low rank representation (LRR) [16, 17] express the samples as a linear combination of other samples. This is expressed as,

$$x_i = X_{i^c} c_i, \forall i \text{ in } \{1,...,n\} \qquad (1)$$

Here $x_i$ ($\in \mathbb{R}$ ) denotes the $i^{th}$ sample and $X_{i^c}$ ($\in \mathbb{R}$ ) all other samples; $c_i$ ($\in \mathbb{R}$ ) is the corresponding linear weight vector.

For all the sub-space clustering techniques the general learning formulation can be expressed as follows,

$$\min_{c_i} \left\| x_i - X_{i^c} c_i \right\|_2^2 + R(c_i), \forall i \text{ in } \{1,...,n\} \qquad (2)$$

Here $R$ is the regularization term. Depending on its nature there are three formulations. For LLMC there is no regularization. For sparse subspace clustering, $R$ is a sparsity promoting $l_1$-norm [13] or $l_0$-norm [14, 15]. For LRR, $R$ is a low-rank penalty usually in the form of nuclear norm.

For each formulation, once the coefficient matrix $C = [\tilde{c}_1, ..., \tilde{c}_n]$ is obtained for all $n$ samples, the affinity matrix is computed. Note that $\tilde{c}_i (\in \mathbb{R})$ is defined from $c_i (\in \mathbb{R})$ by putting zero in the $i^{th}$ position. There is no unique definition to the affinity matrix; the only requirement is that it needs to be symmetric. Several variants have been proposed [4]. For example one option can be –

$$A = |C| + |C^T| \quad (3)$$

This is usually used in SSC.

Another option for LRR is to form the affinity matrix from the scaled left singular values of $C$; this is defined as,

$$A_{ij} = \left(\left[\tilde{U}\tilde{U}^T\right]_{ij}\right) \quad (4)$$

where $\tilde{U}$ and $C = USV^T$.

Yet another way to generate the affinity matrix (usually for LLMC) is by.

$$A = C + C^T - C^T C \quad (5)$$

Once the affinity matrix is defined (by using any suitable formula), one needs to segment the clusters. Usually spectral clustering algorithm (Normalized-Cuts) [18] is used for this purpose.

Related to subspace clustering are the matrix factorization based approaches. Some of them have been referred before. For a review one can peruse [19]. The basic idea there in, is to identify arbitrary subspaces from the data by factoring it.

## 2.2 Transform Learning

Transform learning analyses the data by learning a transform / basis to produce coefficients. Mathematically this is expressed as,

$$TX = Z \quad (6)$$

Here $T$ is the transform, $X$ is the data and $Z$ the corresponding coefficients. The following transform learning formulation was proposed in [4] –

$$\min_{T,Z} \|TX - Z\|_F^2 + \lambda \left(\|T\|_F^2 - \log \det T\right) + \mu \|Z\|_1 \quad (7)$$

The parameters ($\lambda$ and $\mu$) are positive. The factor $-\log \det T$ imposes a full rank on the learned transform; this prevents the degenerate solution ($T=0, Z=0$). The additional penalty $\|T\|_F^2$ is to balance scale.

In [4], an alternating minimization approach was proposed to solve the transform learning problem. This is given by –

$$Z \leftarrow \min_Z \|TX - Z\|_F^2 + \mu \|Z\|_1 \quad (8a)$$

$$T \leftarrow \min_T \|TX - Z\|_F^2 + \lambda \left(\|T\|_F^2 - \log \det T\right) \quad (8b)$$

Updating the coefficients (8a) is straightforward using one step of soft thresholding,

$$Z \leftarrow signum(TX) \cdot \max\left(0, abs(TX) - \mu\right) \quad (9)$$

Here '·' indicates element-wise product.

The update for the transform (8b) also has a closed form solution. This is given as –

$$XX^T + \lambda I = LL^T \quad (10a)$$

$$L^{-1}XZ^T = USV^T \quad (10b)$$

$$T = 0.5U\left(S + (S^2 + 2\lambda I)^{1/2}\right)V^T L^{-1} \quad (10c)$$

The proof for convergence of such an alternating update algorithm can be found in [21].

## 3 Proposed Formulations

We have discussed in the previous sections the existing concepts of subspace clustering and transform learning. Our contribution in this paper is to embed three subspace clustering formulations, associated to three distinct choices for the regularization term R, into the transformed space, that is, instead of learning the affinity matrix from the raw data, we propose to learn it from the (transform) coefficient space.

The trend of learning representations for machine learning has been used rampantly in supervised problems. For example consider [22, 23]. In [22] it is assumed that even if the data cannot be discriminated in the original domain, they can be learnt to be discriminative in the feature domain using dictionary learning. Similarly in [23], it is assumed that eventhough one cannot linearly project the raw samples to their corresponding class labels, the learnt dictionary coefficients can be. Similar ideas are echoed in [24] using the transform learning approach. As mentioned before, similar ideas have been explored for clustering as well [18, 19]. In [25], it has been assumed that even if the data is not clustered on subspaces, its deeply (stacked autoencoder) learnt representation will be. Our work follows the same trend; we assume that even if the original data do not lie on separate subspaces, the (transform) learnt coefficients will lie on different subspaces and hence the ensuing affinity matrix can be segmented for clustering.

A naïve solution would be to learn the transform on the data and then use the coefficients as inputs for subspace clustering. But such a piecemeal formulation will not yield the best results. This can be seen from [25] and [26]. In [26], a deep representation is first learnt and then a third-party clustering algorithm is used on the learnt representation; the results improve in [25] when the deep representation and the clustering are jointly learnt.

We propose to formulate a joint solution instead. Mathematically, our formulation is expressed as,

$$\min_{T,Z,C} \|TX - Z\|_F^2 + \lambda \left(\|T\|_F^2 - \log \det T\right) \\ + \mu \|Z\|_1 + \gamma \sum_i \|z_i - Z_{i^c} c_i\|_2^2 + R(C) \quad (11)$$

Alternating minimization [27] approach is used for solving (11). It can be segregated into the following subproblems.

$$P1: \min_T \|TX - Z\|_F^2 + \lambda \left(\|T\|_F^2 - \log \det T\right)$$

$$P2: \min_Z \|TX - Z\|_F^2 + \mu \|Z\|_1 + \gamma \sum_i \|z_i - Z_{i^c} c_i\|_2^2$$

$$P3: \min_C \sum_i \|z_i - Z_{i^c} c_i\|_2^2 + R(C)$$

The update for the transform (P1) remains the same as in (10). The update for the transform coefficients can also be shown to be the same as in (9). Indeed we have,

$$\min_Z \|TX - Z\|_F^2 + \mu\|Z\|_1 + \gamma\|Z(I-C)\|_F^2$$
$$\Rightarrow \min_Z \|X^T T^T - Z^T\|_F^2 + \mu\|Z\|_1 + \gamma\|(I-C)^T Z^T\|_F^2 \quad (12)$$
$$\Rightarrow \min_Z \left\| \begin{pmatrix} X^T T^T \\ 0 \end{pmatrix} - \begin{pmatrix} I \\ \sqrt{\gamma}(I-C)^T \end{pmatrix} Z^T \right\|_F^2 + \mu\|Z^T\|_1$$

The update for P3 will be dependent on the regularization term $R(C)$. For transformed locally linear manifold clustering (TLLMC) there is no regularization, hence it takes the form –

$$\min_C \sum_i \|z_i - Z_{i^c} c_i\|_2^2 \quad (13)$$

This can be solved for each $c_i$ via the pseudo-inverse.

For transformed sparse subspace clustering (TSSC), the regularization is a sparsity enhancing penalty. In particular if one uses the $l_1$-norm, P3 becomes equivalent to –

$$\min_C \|Z - ZC\| + \mu\|C\|_1 \quad (14)$$

This is a standard $l_1$-minimization for which we have used the iterative soft thresholding algorithm [28].

For transformed low rank representation (TLLR) formulation, a low rank penalty is imposed on $C$. Here we will consider the nuclear norm penalty (defined as the sum of singular values of the matrix), so that P3 reads –

$$\min_C \|Z - ZC\| + \mu\|C\|_* \quad (15)$$

The above problem can be solved using singular value shrinkage [29].

Once the matrix C is obtained, the affinity matrix can be created using either of (3), (4) or (5); here we have retained (3) since it appears to yield the best practical results. Spectral clustering is then applied to the affinity matrix in order to segment the data $X$.

### 3.1 Kernelization

An efficient strategy to handle non-linearity in data in machine learning is to employ the so-called kernel trick. If the (transform) learnt coefficients are not linearly separable into subspaces, we assume that projecting them non-linearly to a higher dimension can make them separable into subspaces. This can be achieved by kernelizing the transform subspace clustering formulation. Kernelizing the basic transform learning has been proposed in [30]. In this work, we incorporate subspace clustering into the kernel transform learning formulation.

In kernel transform learning, a non-linear version of the data is represented in terms of a transform made up of linear combination of non-linear version of itself. This is expressed as,

$$\underbrace{\phantom{(\varphi(X)^T)}}_{transform} (\phantom{\varphi(X)}) = Z \quad (17)$$

One can immediately identify the kernel defined by $K = \varphi(X)^T \varphi(X)$. This allows us to express the kernel version of transform learning in the known (transform learning) form –

$$BK = Z \quad (18)$$

Here, the kernel matrix $K$ plays the role of the data matrix $X$ in the original transform learning formulation; $B$ is the linear weights (similar to transform) that needs to be learnt and $Z$ is the corresponding coefficient matrix.

Under this notation, we can formulate all our versions of transformed subspace clustering using the generic notation –

$$\min_{B,Z,C} \|BK - Z\|_F^2 + \lambda\left(\|B\|_F^2 - \log\det B\right) \\ + \mu\|Z\|_1 + \gamma\sum_i \|z_i - Z_{i^c} c_i\|_2^2 + R(C) \quad (19)$$

The solution for (19) remains the same as in the previous linear case. We can expect improvements over the basic transform learning version with the kernel trick, since the later accounts for non-linearities.

## 4 Experimental Evaluation

### 4.1 Image Clustering

In this section we compare our method with three state-of-the-art deep clustering benchmarks – deep sparse subspace clustering (DSC) [25], deep K-means clustering (DKM) [31] and deep matrix factorization (DMF) [32]. The said studies have been published recently and have compared with traditional clustering techniques like matrix factorization, spectral clustering, subspace clustering, hierarchical clustering etc. Therefore, we do not compare with the traditional ones. We also do not compare with [26], since it has been conclusively shown in [25] that the later excels over the former by a large margin. We have not also compared with recent techniques such as [33], since it is not a pure clustering formulation; it uses label information. We also do not compare against unpublished non-reviewed archival studies.

We follow the experimental protocol from [25, 15]. Experiments were carried out on the COIL20 (object recognition) [34] and Extended YaleB (face recognition) [35] datasets. The COIL20 database contains 1,440 samples distributed over 20 objects, where each image is with the size of 32×32. The used YaleB consists of 2,414 samples from 38 individuals, where each image is with size of 192×168. For both the datasets DSIFT (dense scale invariant feature transform) features were extracted. They were further reduced by PCA to a dimensionality of 300. We experiment on these input features. Since the ground truth (class labels) for these datasets are available, clustering accuracy was measured in terms of Accuracy, NMI (normalized mutual information), ARI (adjusted rand index), Precision and F-score. The results are shown in Table 1 (COIL20) and Table 2 (YaleB). Since the last stage of clustering involves K-means, we ran the experiment 100 times and report the average values.

The parametric settings for the methods compared against have been taken from the respective papers. For our proposed technique, we have kept $\lambda=\mu=0.1$ and $\gamma=1$. TLLMC does not require specification of any other parameter. TSC has $\mu=0.1$ as the sparsity promoting term and TLLR has $\mu=0.01$ as the rank deficiency term. The algorithms are robust to these parametric values; changes by an order of magnitude to either side do not affect the results statistically.

TABLE 1

Comparison showing improvement of proposed variants over the state-of-the-art on COIL 20

| Metric | DSC | DKM | DMF | TLLMC | | TSSC | | TLRR | |
|---|---|---|---|---|---|---|---|---|---|
| | | | | Jt. | PM | Jt. | PM | Jt. | PM |
| Accuracy | .85 | .88 | .86 | **.90** | .88 | **.90** | .88 | .79 | .75 |
| NMI | .91 | .94 | .92 | **.97** | .86 | **.98** | .86 | .89 | .80 |
| ARI | .84 | .86 | .85 | **.86** | .82 | **.88** | .83 | .78 | .71 |
| Precision | .82 | .85 | .84 | **.85** | .78 | **.88** | .79 | .69 | .67 |
| F-measure | .85 | .87 | .84 | **.90** | .84 | **.92** | .86 | .79 | .72 |

*Jt. – Jointly Learnt; PM - piecemeal

TABLE 2

Comparison showing improvement of proposed variants over the state-of-the-art on Yale B

| Metric | DSC | DKM | DMF | TLLMC | | TSSC | | TLRR | |
|---|---|---|---|---|---|---|---|---|---|
| | | | | Jt. | PM | Jt. | PM | Jt. | PM |
| Accuracy | .88 | .91 | .89 | **.98** | .94 | **.98** | .94 | .81 | .79 |
| NMI | .90 | .92 | .90 | **.97** | .95 | **.98** | .94 | .89 | .85 |
| ARI | .83 | .90 | .83 | **.96** | .90 | **.96** | .91 | .73 | .71 |
| Precision | .79 | .91 | .80 | **.97** | .91 | **.98** | .91 | .65 | .65 |
| F-measure | .83 | .90 | .84 | **.92** | .88 | **.95** | .92 | .74 | .71 |

*Jt. – Jointly Learnt; PM - peacemeal

TABLE 3

Comparison of Different Kernels (TLLMC) on proposed variants for COIL 20

| Metric | Linear | Poly-2 | Poly-3 | Poly-4 | Laplacian | Gaussian |
|---|---|---|---|---|---|---|
| Accuracy | .90 | .81 | .80 | .72 | .90 | **.93** |
| NMI | .97 | .91 | .89 | .81 | .95 | **.97** |
| ARI | .86 | .81 | .78 | .80 | .86 | **.87** |
| Precision | .85 | .80 | .77 | .76 | .84 | **.85** |
| F-measure | .90 | .83 | .80 | .82 | .89 | **.90** |

TABLE 4

Comparison of Different Kernels (TLLMC) on proposed variants for YALE B

| Metric | Linear | Poly-2 | Poly-3 | Poly-4 | Laplacian | Gaussian |
|---|---|---|---|---|---|---|
| Accuracy | .98 | .92 | .91 | .88 | .98 | **.98** |
| NMI | .97 | .87 | .95 | .83 | .96 | **.97** |
| ARI | .96 | .95 | .88 | .90 | .96 | **.96** |
| Precision | .97 | .90 | .87 | .87 | .97 | **.98** |
| F-measure | .92 | .86 | .85 | .81 | .92 | **.92** |

TABLE 5

Comparison of Different Kernels (TSSC) on proposed variants for COIL 20

| Metric | Linear | Poly-2 | Poly-3 | Poly-4 | Laplacian | Gaussian |
|---|---|---|---|---|---|---|
| Accuracy | .90 | .96 | .96 | .94 | .98 | **.98.** |
| NMI | .98 | .90 | .88 | .84 | .93 | **.94** |
| ARI | .88 | .87 | .87 | .83 | .92 | **.94** |
| Precision | .88 | .84 | .83 | .81 | **.89** | .88 |
| F-measure | .92 | .82 | .82. | .80 | **.89** | .88 |

TABLE 6

Comparison of Different Kernels (TSSC) on proposed variants for YALE B

| Metric | Linear | Poly-2 | Poly-3 | Poly-4 | Laplacian | Gaussian |
|---|---|---|---|---|---|---|
| Accuracy | .98 | .76 | .76 | .74 | .99 | **.99** |
| NMI | .98 | .81 | .83 | .80 | .96 | **.98** |
| ARI | .96 | .83 | .84 | .81 | .95 | **.96** |
| Precision | .98 | .88 | .89 | .89 | .99 | **.99** |
| F-measure | .95 | .81 | .83 | .82 | .95 | **.96** |

In Tables 1 and 2, we compare our proposed techniques TLLMC, TSSC and TLLR with DSC, DKM and DMF. Our proposed version comes in two flavors. In the joint (Jt.) formulation (proposed in this work), SSC is embedded in the transform learning; this is the one proposed in this paper. The second formulation is the simplistic piecemeal (PM) formulation. Here features obtained by transform learning are fed into subspace clustering in a piecemeal fashion; the two (transform learning and clustering) are not learnt jointly. As one can expect, the joint formulation yields much better results than the piecemeal formulation. This has been seen in prior studies as well. In [25], the clustering formulation was embedded in a deep autoencoder, where as in [26] the formulation was piecemeal – the deep autoencoder was learnt separately and the features from the bottleneck layer fed into a separate clustering algorithm. It was shown in [26] that the joint formulation yields much better results than the piecemeal [26] formulation.

From Tables 1 and 2, we also observe that the TLLR formulation does not yield good results. This is not surprising; it follows from the findings in [25]; there in it was find that LRR based formulations yield poor results on these datasets. Our joint TSSC and TLLMC formulations always yields the best results.

Next we show the empirical convergence plot. The plots are generated for the COIL20 dataset. On the X-axis is the iteration number and on the Y-axis is the normalized cost function. All the plots show that our algorithm converges very fast, in less than 10 iterations.

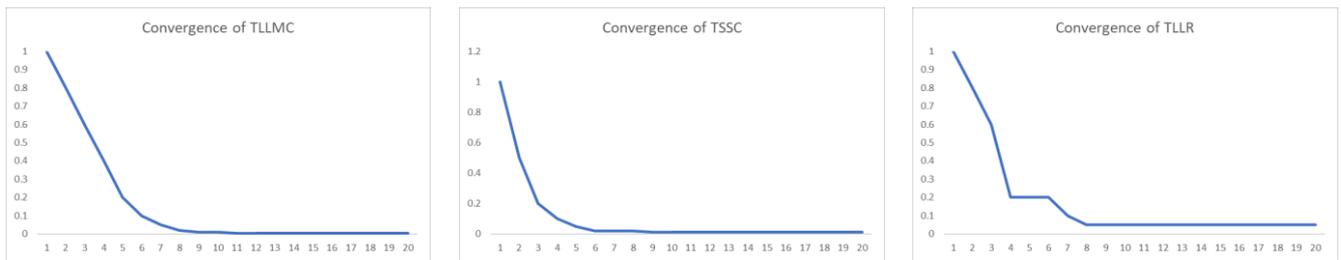

Figure 1. Empirical Convergence Plot for Image Clustering

All the aforesaid results are shown for linear kernels. We have discussed in III.A how our method can be kernelized. In Tables 3 to 6, we show results of kernelized versions for COIL 20 and Yale B respectively. Note that these results are shown for the jointly learnt variants since they yield better results than their corresponding piecemeal versions. In Tables 3 and 4, we show the results for kernelized variants of TLLMC and in Tables 5 and 6, we show results for kernelized variants of TSSC. We do not show the results for TLLR since it performs poorly.

The results are shown for some typical kernels like linear, Gaussian, Laplacian and polynomial kernels of order 2, 3 and 4. The RBF kernel yields the best results almost always; followed by the Laplacian kernel. The polynomial kernels yield the worst results consistently. They are worse than the simple linear technique; the results deteriorate with the increase in the order of the polynomial.

All the experiments were run on an Intel i7 processor with 32 GB RAM running a 64 bit Windows 10. The proposed techniques and DMF was based on Matlab; DSC, DKM were based on Python. The run-times are shown in Table 5.

TABLE 5
Comparison of Runtime in Seconds

| Technique | Coil 20 | Yale B |
|---|---|---|
| DSC | 62 | 61 |
| DKM | 87 | 83 |
| DMF | 57 | 54 |
| TLLMC | 11 | 9 |
| TLLMC-Kernel | 44 | 38 |
| TSSC | 12 | 11 |
| TSSC-Kernel | 50 | 42 |
| TLRR | 16 | 13 |
| TLRR-Kernel | 58 | 48 |

The runtimes show that our methods are the fastest. The linear versions are obviously faster than the kernelized ones. This is because the size of the kernelized data matrix is much larger than the original data matrix (input to the linear versions). Of the proposed three variants, TLLMC is the fastest as it has no regularization term. TSSC is slightly slower owing to the requirement of the thresholding step in every iteration. TLRR is even slower, because one needs to threshold the singular values; and computing singular values in every iteration is time consuming. All the compared deep learning based techniques are slow; this is because they need to learn multiple layers.

## 4.2 Document Clustering

Our second example focuses on document clustering. For this application, we follow the protocol defined in a recent work [36]. For our experiments, we use three data sets TDT2 corpus [37], Reuters-21578 corpus [37], and 20 Newsgroup [37].

The TDT2 English document data set includes six months of material drawn on a daily basis from six English language news sources. In this set, the total number of samples is 9394, the feature dimension is 36771, and the number of clusters is 30.

The Reuters-21578 document set is a collection of manually categorized newswire stories from Reuters Ltd. In this set, the total number of samples is 8293, the feature dimension is 18933, and the number of clusters is 65.

The 20 Newsgroups data set is a collection of approximately 20,000 newsgroup documents. In this set, the total number of samples is 18846, the feature dimension is 26214, and the number of clusters is 20.

Following [36], we report the result in terms of two metrics namely entropy and purity. For good clustering, one requires small entropy and high purity [38]. In [36], the metrics are reported by varying the number of clusters from 2 to 10. We follow the same protocol.

Suppose there is ground truth data that labels the samples by one of classes. Purity is given by,

$$purity = \frac{1}{n}\sum_{k=1}^{r} \max_{1 \le l \le q} n_k^l \quad (20)$$

where $n_k^l$ is the number of samples in the cluster $k$ that belong to original class $l$. A larger purity value indicates better clustering performance.

Entropy measures how classes are distributed on various clusters. The entropy of the entire clustering solution is computed as,

$$entropy = \frac{1}{n \log_2 q}\sum_{k=1}^{r}\sum_{l=1}^{q} n_k^l \log_2 \frac{n_k^l}{n_k} \quad (21)$$

where $n_k = \sum_l n_k^l$. Generally, a smaller entropy value corresponds to a better clustering quality

Our proposed method has been compared with [36] and with [39]; to the best of our knowledge these are two of the most recent works in this area and are known to yield the best possible results on document clustering.

We found that the kernelized versions yield better results than the linear counterparts; and the best results are obtained from the Gaussian kernel. These results are shown here.

TABLE 6
Comparison showing improvement of proposed method over existing techniques on TDT2

| Clusters | Entropy (lower is better) | | | | | Purity (higher is better) | | | | |
|---|---|---|---|---|---|---|---|---|---|---|
| | CFAN | SNMF-PCA | TLLMC | TSSC | TLLR | CFAN | SNMF-PCA | TLLMC | TSSC | TLLR |
| 2 | .0000 | .0000 | .0000 | **.0000** | .0000 | 1.0000 | 1.0000 | 1.0000 | **1.0000** | 1.0000 |
| 4 | .0000 | .0159 | .0000 | **.0000** | .0159 | 1.0000 | .9956 | 1.0000 | **1.0000** | .9229 |
| 6 | .0526 | .0889 | .0063 | **.0013** | .0431 | .9435 | .8954 | .9481 | **.9963** | .8917 |
| 8 | .0552 | .0941 | .0329 | **.0465** | .0506 | .9476 | .8801 | .9490 | **.9013** | .9661 |
| 10 | .0808 | .0685 | .0188 | **.0178** | .0631 | .9153 | .9224 | .9265 | **.9775** | .9254 |
| Avg. | .0312 | .0582 | .0126 | **.0103** | .0682 | .9682 | .9285 | .9673 | **.9736** | .9527 |

TABLE 7
Comparison showing improvement of proposed method over existing techniques on REUTERS

| Clusters | Entropy (lower is better) | | | | | Purity (higher is better) | | | | |
|---|---|---|---|---|---|---|---|---|---|---|
| | CFAN | SNMF-PCA | TLLMC | TSSC | TLLR | CFAN | SNMF-PCA | TLLMC | TSSC | TLLR |
| 2 | .0651 | .0651 | .0651 | **.0493** | .0651 | .9912 | .9912 | **.9912** | .9735 | **.9912** |
| 4 | .3751 | .3700 | .3257 | **.2103** | .2508 | .7934 | .8035 | .8618 | **.8984** | .8061 |
| 6 | .2029 | .2521 | .3268 | **.1905** | .2435 | .8719 | .8380 | .8040 | **.8855** | .8595 |
| 8 | .2258 | .2829 | .1646 | **.2811** | .2279 | .8586 | .8076 | .9252 | **.9135** | .8524 |
| 10 | .3677 | .4464 | .3166 | **.2579** | .3286 | .6690 | .6288 | .7586 | **.8069** | .7331 |
| Avg. | .2360 | .2701 | .2213 | **.1978** | .2334 | .8502 | .8260 | .8777 | **.8950** | .8605 |

TABLE 8
Comparison showing improvement of proposed method over existing techniques on NewsGroup

| Clusters | Entropy (lower is better) | | | | | Purity (higher is better) | | | | |
|---|---|---|---|---|---|---|---|---|---|---|
| | CFAN | SNMF-PCA | TLLMC | TSSC | TLLR | CFAN | SNMF-PCA | TLLMC | TSSC | TLLR |
| 2 | .8556 | .9843 | .8841 | **.8172** | .8131 | .6867 | .5500 | .6500 | **.7233** | .7233 |
| 4 | .6065 | .7301 | .6114 | **.5911** | .5755 | .6083 | .5183 | .6540 | **.6567** | .6575 |
| 6 | .5441 | .6492 | .4835 | **.4697** | .5711 | .6017 | .5322 | .6806 | **.7050** | .6028 |
| 8 | .5505 | .6562 | .4989 | **.4673** | .5942 | .5708 | .4700 | .6338 | **.6721** | .5259 |
| 10 | .5395 | .6025 | .4602 | **.4449** | .5399 | .5543 | .4977 | .6433 | **.6690** | .5600 |
| Avg. | .6037 | .7010 | .5664 | **.5490** | .6120 | .6002 | .5176 | .6554 | **.6793** | .6013 |

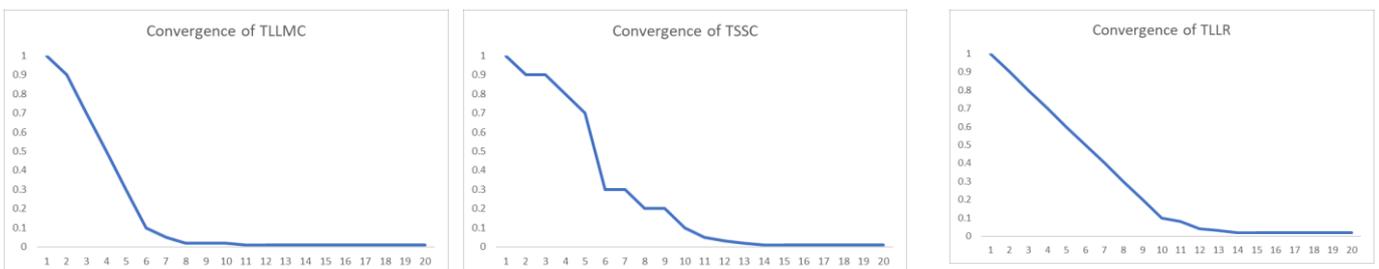

Figure 2. Empirical Convergence for Document Clustering

The results show that only for the easiest dataset (TDT2), CFAN yields slightly better results than some of our proposed variants. Note that, by easiest we mean the dataset where all the algorithms yield good results. But even for this dataset the best result is obtained from TSSC. For the more difficult datasets, all our proposed methods excel over the ones compared against; that too by a significant margin.

We show the empirical convergence plot of our proposed algorithms in Figure 1. The results are shown for the Reuters dataset with 10 clusters. The X-axis shows the iteration numbers and the Y-axis the normalized objective function. By normalized, we mean the cost at the current iterate divided by the initial cost. The plots show that all our algorithms converge very fast – within 20 iterations. The results for the other datasets are similar; given the limitations of space, we do not show them here.

Finally we show the runtimes for all the techniques. The configuration of the machine remains the same as before. All the algorithms have been implemented in Matlab.

TABLE 5
Comparison of Runtime in Seconds

| Technique | TDT2 | Reuters | NewsGroup |
|---|---|---|---|
| CFAN | 308 | 221 | 682 |
| SNMF-PCA | 276 | 195 | 349 |
| TLLMC | 482 | 303 | 794 |
| TSSC | 507 | 346 | 733 |
| TLLR | 793 | 515 | 901 |

The timing comparison shows that our proposed methods are slower than the benchmarks. This is because, unlike image clustering, the state-of-the-art in document clustering are based on matrix factorization and not deep learning. Matrix factorization is a matured area with fast algorithms. Similar to matrix factorization, our proposed techniques are all shallow. However, we require computing SVDs for the updating the transform in every iteration. This is time consuming and hence is slower than matrix factorization.

## 5 Conclusion

In this work, we have incorporated subspace clustering formulations into the transform learning framework. This results in three variants – transformed locally linear manifold clustering, transformed sparse subspace clustering and transformed low rank representation. We have also propose kernelized versions of the aforesaid three variants.

Experiments have been carried out on two benchmark problems – image and document clustering. For each problem, state-of-the-art techniques are compared against. In all cases, our proposed method excels over these in terms of the metrics used here.

In future, we would like to build on the proposed approach in two ways. One way will be to try to incorporate other clustering cost functions. For example in [31], it has been shown how K-means can be embedded into deep autoencoders. We would like to explore a similar avenue.

The other enhancement to this work would be to try to make the proposed approach deeper. In a recent study, the concept of deep transform learning has been proposed [40]. We would like to see if it would be possible to embed clustering formulations into the said deep formulation.

## Acknowledgement

The first author is partially supported by TCS Research. All the authors are supported by the Indo-French CEFIPRA grant DST-CNRS-2016-02.